# Explaining human body responses in random vibration: Effect of motion direction, sitting posture, and anthropometry

M. M. Cvetković, R. Desai, K. N. de Winkel, G. Papaioannou, and R. Happee

*Abstract* — This study investigates the effects of anthropometric attributes, biological sex, and posture on translational body kinematic responses in translational vibrations. In total, 35 participants were recruited. Perturbations were applied on a standard car seat using a motion-based platform with 0.1 to 12.0 Hz random noise signals, with 0.3 m/s$^2$ rms acceleration, for 60 seconds. Multiple linear regression models (three basic models and one advanced model, including interactions between predictors) were created to determine the most influential predictors of peak translational gains in the frequency domain per body segment (pelvis, trunk and head). The models introduced experimentally manipulated factors (motion direction, posture, measured anthropometric attributes, and biological sex) as predictors. Effects of included predictors on the model fit were estimated. Basic linear regression models could explain over 70% of peak body segments' kinematic body response (where the R$^2$ adjusted was 0.728). The inclusion of additional predictors (posture, body height and weight, and biological sex) did enhance the model fit, but not significantly (R$^2$ adjusted was 0.730). The multiple stepwise linear regression, including interactions between predictors, accounted for the data well with an adjusted R$^2$ of 0.907. The present study shows that perturbation direction and body segment kinematics are crucial factors influencing peak translational gains. Besides the body segments' response, perturbation direction was the strongest predictor. Adopted postures and biological sex do not significantly affect kinematic responses.

## I. INTRODUCTION

Understanding the effects of vibration on human health and performance is essential for motion comfort. To comprehend those impacts, it is necessary to determine the interactions effects between potential predictors of kinematic body responses [1], [2]. Although vibrations in real-world situations are frequently multi-axial, the dominance of particular directions depends on the vibration environment [1]. To gain knowledge of how the human body responds to a demanding driving condition, where the perturbation bandwidth is wider, it is necessary to assess interactions between vibrational characteristics [3], support of the body (seat design, back support, and inclinational angle), postures, [4]–[6] anthropometric characteristics [4], amongst others. Understanding any interactions between those variables, especially in multi-axial perturbation, could ultimately increase motion comfort. A number of studies investigates human body responses in vertical direction [7], [8], while simple [9] and advanced human biomechanical models [10] were used to further explore and predict human body responses in multi-axial perturbation. Wider inter-model variability can be observed within the advanced biomechanical models, which may yield more insightful conclusions.

Anthropometric characteristics, especially body weight, are important when determining whole-body vibrations [11]. Mainly, more considerable body weight, and in particular a higher percentage of body mass in the buttocks-thighs, causes greater contact between the seated person and the seat [12], which could considerably alter 'seat-to-the-body' biodynamic responses [13]. Additional factors, such as seat back position [6] and inclinational angle [3] also affect kinematic body responses. For instance, the resonance frequencies of biodynamic responses increase as the backrest inclination increases (from 30$^0$ to 90$^0$) [3].

It should be noted that discomfort develops over prolonged driving. The driving discomfort can influence adopted sitting strategies, which will change over time [14]. Alternating the adopted sitting posture will direct how the body responds to motion direction and its characteristics. Non-linearity in biomechanical responses in tri-axial perturbation can be seen when comparing slouched and upright active sitting postures [6]. A slouched posture will result in larger trunk translational responses but smaller head responses when compared to an erect posture.

In the present study, we take into consideration interaction effects between participants' anthropometric attributes and experimentally manipulated factors (sitting postures and tri-axial motion) on kinematic body responses. We explore and explain the effects on the pelvis, trunk, and head peak translational gains. To achieve the study objectives, test participants were subjected to randomized vibration in fore-aft, lateral, and vertical directions, adopting a different sitting position (e.g., upright active, upright passive, and slouched). A standard car seat was used to explore how such a *"one size fits all"* design accommodates a wide population. Exploratory linear regression models were created to examine the kinematic body response considering posture, motional direction, and measured anthropometric attributes. One of the specific study objectives is to determine the importance of the postural alternation on the body segments' kinematic responses. A subsequent stepwise regression analysis was conducted to establish the main and interaction effects between personal characteristics and experimentally manipulated variables.

## II. METHODOLOGY

The experiment was based on the authors' previous work [6] where they designed an experimental protocol to determine how the transmission of vibrations in seated occupants is affected by seat back height and posture.

*1) Participants*

The experiment was conducted with thirty-five healthy participants, of which 19 were male and 16 females, with an average age (± standard deviation) of 33.8 (± 11.5) years, a height of 174.2 (± 9.1) cm, and a weight of 73.3 (± 15.4) kg. An informational letter was provided to the participants listing the benefits and the possible risks of participation. Candidates with existing pain/discomfort, present in the last seven days,

Authors are with the Faculty of Mechanical, Maritime and Materials Engineering, Delft University of Technology, 2628 CD Delft, The Netherlands (e-mail: m.cvetkovic-3@tudelft.nl).

or musculoskeletal disorders, which could conceivably influence the quality of collected data, were excluded from the experiment. During the trials, participants' well-being was constantly monitored, and in case of reporting severe motion sickness symptoms, through the documentation of higher misery scores [15], the process was terminated and collected data were omitted from the analysis. Each participant's contribution to the data collection was reimbursed with a 20 EUR gift card.

*2) Ethical statements*

Keeping in mind the Helsinki Declaration, written consent, data collection procedures, and respect for participant privacy were all adhered to in order to ensure the research was ethically acceptable. The procedure was recognized and approved by the Human Research Ethics Committee of the Delft University of Technology (HREC), under application number 962.

### B. Driving simulator

The data were collected at the Delft University of Technology using an Advanced Vehicle Simulator (DAVSi), Figure 1. The DAVSi is a multiaxial device (hexapod), on which a mock-up half-car Toyota Yaris is placed.

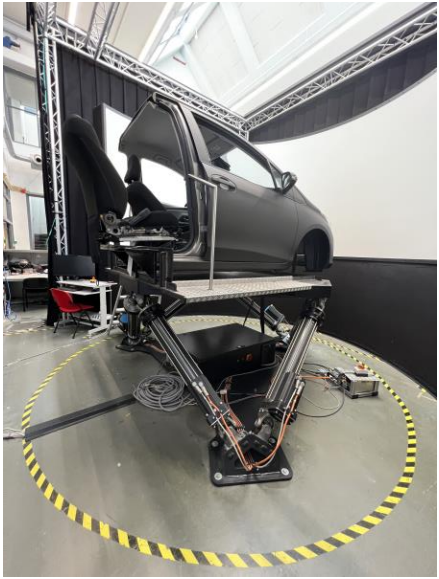

Figure 1. DAVSi (Delft Advanced Vehicle Simulator)

*1) Input vibrations*

To excite the motion platform, a random noise signal with a frequency bandwidth from 0.1 to 12.0 Hz, with 0.3 m/s$^2$ rms power, was generated (Figure 2). The signal was adopted to simulate comfortable driving conditions, providing also good coherence levels between simulated platform motion and measured body segment's (i.e., pelvis, trunk, and head) kinematic responses [6]. The signal's frequency range was selected to capture the frequencies that are essential to driving comfort.

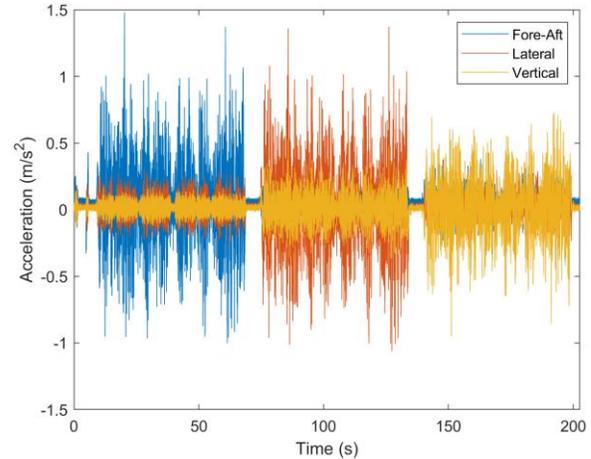

Figure 2. Recorded random vibration in fore-aft, lateral, and vertical direction

Three translational motions were consecutively applied in fore-aft, lateral, and vertical directions (Figure 2). Each directional motion (fore-aft, lateral, and vertical) lasted 60 seconds, with three seconds fade-in and three seconds fade-out to avoid exposing participants to unnecessarily jerky motion. The duration of each trial was 200 seconds, which was repeated for each posture condition.

### C. Human Body Kinematics and adopted sitting strategies

*1) Capturing kinematic body responses*

A tri-axial full-body motion capturing system (MTW Awinda, Xsens Technologies, Enschede, The Netherlands) was used to monitor posture during experiments and guide participants to adopt the desired posture. It was also used to record translational body kinematics and calculate the transmissibility of whole-body vibration from the motion platform to the pelvis, trunk, and head.

*2) Sitting postures*

While being exposed to random vibration in three axial directions, participants were tutored to maintain three sitting postures: passive upright (preferred), active upright (erect), and slouched. Figure 3 presents adopted pelvis-thorax and head-thorax angles during the experimental trials. The angles are defined between the horizontal plane and the connecting line between the pelvis-thorax and cervical spine-head segments. Adopted sitting postures were monitored using the *Xsens* software.

The sitting postures are defined as follows:

- *Passive upright (preferred)*: For this strategy, the participants placed their buttocks in the middle of the seat, leaned their body on the seat back, adopted extended knee angles, placed their upper limbs in the lap and relaxed them. This strategy aimed to achieve relaxed upper body muscles while still performing an erect posture.

- *Active upright (erect):* For this strategy, the participants placed the buttocks at the most posterior position on the seat, pressing the abdomen and shoulders outwards, and keeping the back in an arched position. Their upper arms stayed relaxed in the lap and participants extended their knee angles similar to the passive erect posture. This strategy was more demanding compared with the previous, requiring active abdominal and lower/upper back muscles while keeping the S-shape back curvature.

- *Slouched:* For this strategy, the participants moved their pelvis to the posterior seat pan position, resulting directly in flexing the lumbar spine toward C-shape curvature. Upper and lower arms stayed the same as previously described while keeping their chest and head straight.

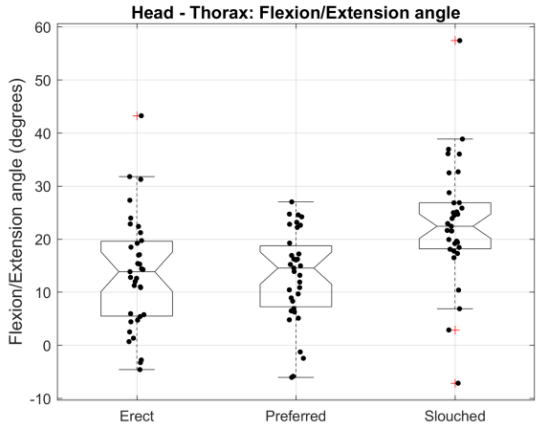

*a) Head – Cervical spine flexion / extension angle*

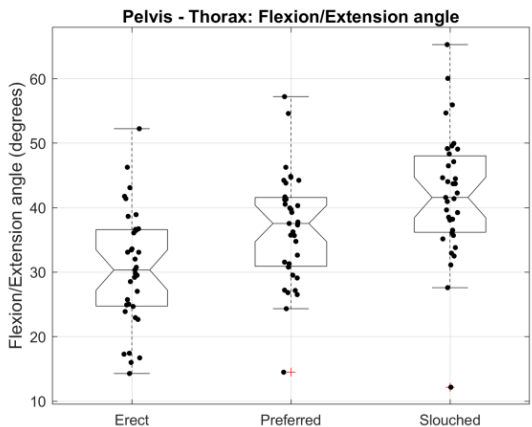

*b) Lumbar spine – Pelvis flexion / extension angle*

Figure 3. Flexion – Extension angles between the body segments

### D. Procedure

The primary goal of the experimental protocol was to ensure good quality of the collected data. Therefore, participants were asked to carefully read the provided information and to get familiar with the procedure (duration, benefits, possible risks, etc.). After agreeing to proceed, the participants signed an informed consent form.

Prior to conducting the experiment, the motion tracking suit was calibratedaccording to a required procedure. Firstly, both body lengths and circumferences were measured. The gathered anthropometry data were used by the XSENS software to reconstruct body segment orientation in three-dimensional space. The model segment calibration was required to align motion trackers with the test subject's anthropometric attributes. Finally, the calibration required the execution of gait motions at a normal pace (back and forth) and then maintaining a standing N pose at the initial position. After the calibration of the motion tracking suit, participants were carefully guided to the top of the motion platform and positioned at the experimental seat which was a standard passenger seat as used in the Toyota Yaris compact car.

Participants were asked to adjust horizontal seat distances to their comfortable convictions. However, seat inclination angle (110º) and vertical distances remained at the initial position. The seat belt was not fastened, to not constrain the body's kinematics. The order of sitting strategies was randomly assigned, and participants were allowed to take a break after each trial. Thereby, development of discomfort or drowsiness caused by the platform motions was mitigated.

### E. Data analysis

#### 1) Calculating body segment gain responses

Transfer functions were calculated to estimate the transmission of the platform motion to the participant's body segments in linear acceleration (fore-aft, lateral, and vertical). This study calculated pelvis, trunk, and head translational gains. The translational gain represents the ratio of the acceleration response of a specific body segment (i.e., pelvis, trunk, and head) in the time domain to the corresponding input vibration (i.e., platform motion) in the time domain. The transfer functions were calculated using Welch's averaged method within Matlab Signal Processing Toolbox (R2021b). The modified periodogram method selected the time signal in 15 segments (windows size 24 seconds, for instance) with 50 percent overlap to reduce variance. The successive blocks were windowed for each section with a Hamming window, and 15 segments of modified periodograms were averaged.

#### 2) Peak translational gains

The translational peak gains were identified to study the effect of vibrational transmissibility on kinematic body responses. To exclude outliers with unexpected body responses, the identification of the peaks was limited to different frequency ranges for the different segments (i.e., pelvis, trunk, and head) depending on the platform motion (i.e., fore-aft, lateral, and vertical). The effect of adopted sitting strategies on body kinematics was further examined by selecting peak translational responses. The frequency range was determined based on the consistency of the previous work [6].

- For fore-aft perturbation to fore-aft translation, peak gains were constrained between 1 and 6 Hz. However, the trunk and pelvis gain responses were selected between 1 and 5.1 Hz and 1.5 and 5.7 Hz, respectively.

- In the case of lateral perturbation, the translational body responses were constrained to the same frequency range (from 1 to 4.3 Hz).

- Vertical platform motion was considered to have a more comprehensive frequency range, between 1 and 7.2 Hz, for all body parts.

According to the data, only 36 out of 918 body segment peak gains exhibited peak translational gains outside the above predefined frequency range.

*F. Statistical Analysis*

A multiple linear regression analysis was performed to explain kinematic body responses while being exposed to fore-aft, lateral, and vertical platform motion.

*1) Basic models – Linear regression models*

Three models were created to explain peak translational segment body responses in tri-axial perturbation. Models were generated in *Matlab*, using the *fitlm* function (Matlab Statistics and Machine Learning Toolbox). The first basic model included experimentally manipulated factors (motion direction and body segments). The predictors were classified as categorical. The responses were observed for the pelvis, trunk, and head in fore-aft, lateral, and vertical directions. The mathematical linear regression model ($mdl_0$) was defined as:

$$mdl_0: gain = \beta_0 + \beta_1 MD + \beta_2 BS \quad (1)$$

where $\beta$ represents the model estimated coefficients; MD defines the perturbation direction (i.e., fore-aft, lateral, and vertical); and BS are body segments (i.e., head, trunk, and pelvis).

The second model ($mdl_1$) further investigated whether the adopted sitting postures (P) (i.e., erect, preferred, and slouched) affect the model prediction.

$$mdl_1: gain = \beta_0 + \beta_1 MD + \beta_2 BS + \beta_3 P \quad (2)$$

Lastly, the third model ($mdl_2$) examines the effects of general body characteristics (Body Weight: BW; Body Height: BH) and biological sex (BS):

$$mdl_2: gain = \beta_0 + \beta_1 MD + \beta_2 BS + \beta_3 P + \beta_4 BW + \beta_5 BH + \beta_6 BS \quad (3)$$

Measured body characteristics (BW and BH) were used as continuous predictors, while BS was a categorical variable.

*2) Exploratory model – Stepwise multiple linear regression*

In addition to experimentally manipulated variables, we investigated effects of subjects' age, anthropometric lengths (body, trunk, and hip height), hip width, and body weight. The remaining body attributes (i.e., segment masses and measured body circumferences) were not used as predictors in the models because of multicollinearity (i.e., high correlation with other selected predicting variables). Besides the anthropometric-based variables, the model included categorical predictors such as biological sex and adopted sitting postures, motion direction, and body segments (i.e., head, trunk, and pelvis).

The *stepwiselm* function of the Matlab Statistics and Machine Learning Toolbox was used to process the data. We included main effects (first order) and second order interaction terms. A criterion in stepwise models were established as "*PEnter*" of 0.01, and "*PRemove*" of 0.05. To identify outliers in the data, Cook's distance was calculated. Data points three times larger than the average were considered outliers and excluded from the regression model. In total, 61 data points out of 918 were classified as outliers.

*3) Estimating the strength of the predictor effect*

The stepwise multiple linear regression selects a set of predictors with statistically significant effects. Additional analysis was conducted to determine the effect size (magnitude) of each predictor on the response (created model). The *Matlab* function *plotEffects* was performed to calculate the main effect size. The function assesses the main effect by representing the impact of one predictor on the model by varying the selected predictor value while averaging out the effects of the other predictors. The interaction effects between the predictors were examined using the *plotInteraction* function (*Matlab*).

## III. RESULTS

*A. Translational peak gain responses*

Thirty-five participants completed the experiment. Due to insufficient and deviant kinematic data (Figure 4), participant 35 (male; age= 26 years; weight=81 kg; body height = 175 cm) was excluded from further analysis. Data corresponding to specific translational peak gains were omitted from the data set if they were outside the predefined frequency range. The remaining participants' data were retained, considering other conditions (i.e., sitting posture, perturbation direction, translational gains). Around 96% of peak transitional gains matched these criteria. The available samples for each signal are illustrated in Figure 4.

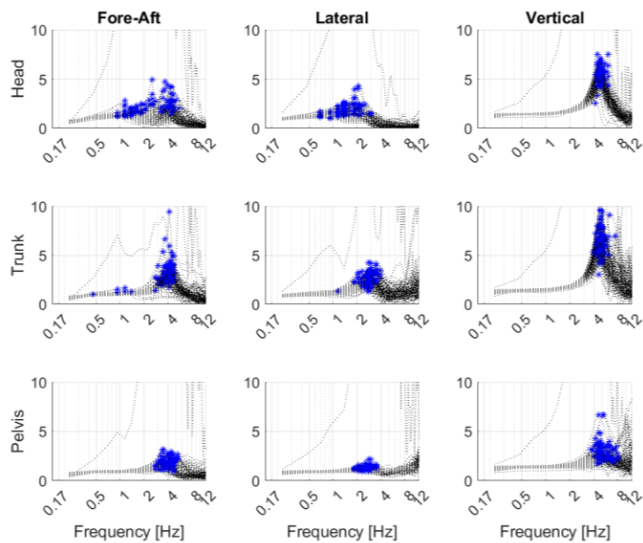

Figure 4. Peak translational gain response of the body segments

*B. Regression models*

Linear regression models were created to investigate the effect of experimentally manipulated factors (posture, motional direction, and defined body segments), measured anthropometric attributes, and biological sex on translational peak body kinematic responses. The models identify the most important variables to explain postural stabilization. In total, 4 models were created, 3 linear (experimentally manipulated variables) and one exploratory model (stepwise multiple linear regression).

*1) Base model: Linear regression models*

The basic regression model ($mdl_0$) of postural stabilization data yielded an overall $R^2$ adjusted (adj.) of 0.728 (F-statistics vs. constant model: 590, p-value <0.001). Almost three-quarters of the variance in translational peak gain of the body responses could be explained by simulator motion direction

(median β = 1.09, and Interquartile Range, IQR = 2.97) and body segment responses (median β = 1.64, IQR = 0.90). The largest estimated coefficients were reported for trunk (β = 2.09) and vertical direction (β = 2.58).The inclusion of posture in the model (mdl$_1$) did not have any significant influence on the coefficient of determination ($R^2$ adj. = 0.730), but the number of predictors within the equation were expanded. Furthermore, the model (mdl$_1$) used all categorical variables to explain the body segments' peak translational gain responses, with only two factors being excluded from the linear model (preferred sitting posture and pelvis). The third, final linear model (mdl$_2$), yielded an $R^2$ adj. of 0.732 ($R^2$ = 0.735, F-statistics vs. constant model: 590, p-value <0.001). As a result, the model included in total most of the predictors (in total eight predictors, with a median estimated coefficient of 0.160, IQR = 1.42).

*2) Exploratory model: Multiple linear regression model*

The exploratory model the predictors included in mdl$_2$, adding measured body length data and participants age as new predictors. The stepwise procedure also introduced interaction terms between predictors, resulting in an improved $R^2$ adj. (0.833) compared to all three basic models. After applying the Cook's method to eliminate the outliers and reduce the predicting models' uncertainty, the $R^2$ adj. further improved and yielded an adjusted coefficient of determination of 0.907 ($R^2$ = 0.914, F-statistic vs. constant model: 137, p-value <0.001).

The exploratory model revealed that gains highly depend on the directional perturbation (vertical perturbation, β = 4.13), body segments' (trunk, β = 4.57), and biological sex (female, β = 12.86). Unexpectedly, the model did not show a significant effect of adopted sitting posture, but rather a weak influence (preferred β = 0.90, and slouched β = 0.67). In addition, the model did not take into account the erect posture (β = 0) when explaining the kinematic body responses. Before being included in the interaction terms, the remaining main predictors did not report an excessive estimated coefficient (median β of remaining predictors was 0.20, with the IQR = 0.96). The fitted model indicated that the measured anthropometric attributes (body height β = 0.68, and hip height β = 0.74) and total body weight (β = 0.31) positively influence the peak translational body segment gains. In contrast, trunk height and hip width are negatively associated, with estimated coefficients of -0.95 and -0.28, respectively.

The stepwise procedure included 46 various interactions between predicting variables. The interaction between platform motion direction and body segments, had the largest coefficient (vertical direction and trunk β =2.48, and vertical direction and head β = 2.44). An interaction was found between biological sex and posture (β = 0.37). Specifically, female participants tend to have larger trunk kinematic responses in the fore-aft, lateral, and vertical perturbation. This was emphasized when the driving simulator was performing vertical motion (β = 0.33).

Adopting a slouched sitting posture while the platform was performing lateral motion, the participants gained more significant body segments' translational gains (β = 0.22). Adopting preferred sitting strategies helped participants reduce kinematic body responses (β = -0.04). Vertical perturbation to vertical translational responses positively interacted with sitting postures (preferred β = 0.03 and slouched β = 0.62).

Details of estimated coefficients and significances are shown in the appendix, available as supplementary material.

*C. Main and interaction effect of model predictors*

*1) Main and interaction effect*

The independent variables' estimated effects on the gains are visualized in Figure 5. The presented results show the effect of individual predictors on the model response by changing the predictor values (from minimum to maximum value) while averaging out the effects of the other predictors.

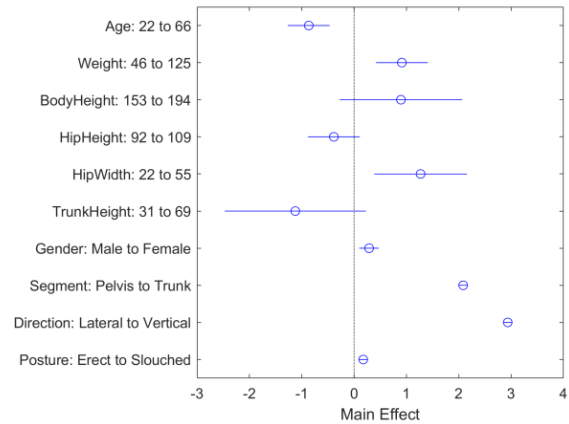

Figure 5. The effect of the main predictors on the model

From Figure 5 it is apparent that the strongest positive effects are the motion direction (from lateral to vertical median is 2.93) and the body segments (from the pelvis to trunk median is 2.09). Additionally, the narrow error bars reflect a low degree of uncertainty. Participants with larger body weight (median of 0.95), stature (median of 1.02), and wider hip with (median of 1.30) will have larger kinematic responses. Those effects come with broader standard errors. Biological sex and adopted sitting strategies demonstrate a relatively smaller positive effect (median of 0.29 and 0.08, respectively). Peak translational gains decrease with increasing participant age (median of -0.89), sitting height or trunk height (median of -1.22), and hip height (median of -0.44, with the largest uncertainty ranging from -2.57 to 0.13).

*2) Interaction effect*

Figure 6 shows the interaction effect between body height and weight (a) and between body height and motion direction (b). The figure shows the conditional effect, where the body height is taken as three distinct fixed values (maximum, average, and minimum value of the calculated body height).

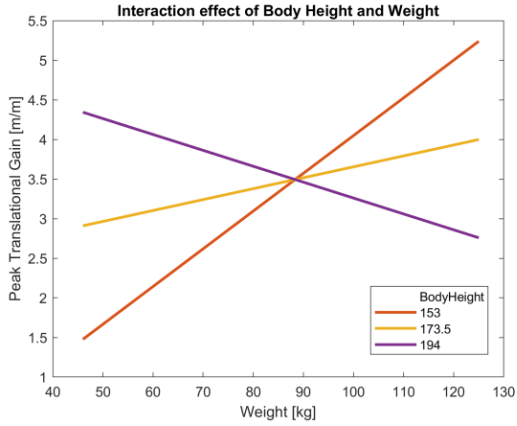

a) Interaction effect between body height and body weight

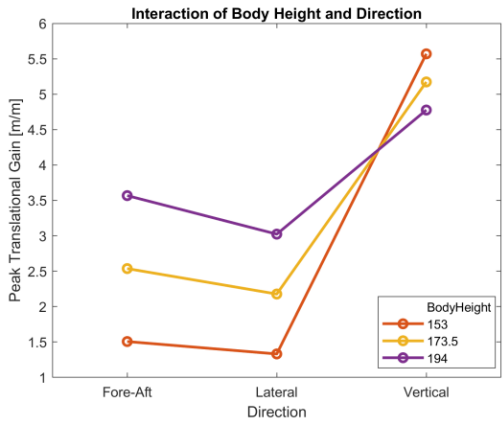

b) Interaction effect between body weight and perturbation direction

Figure 6. Influence of the interaction effect between body height and weight, and body height and tri-axial perturbation on the body segments' peak translational gains.

Body length and weight significantly affect the translational gains (Figure 6, a). A similar interaction was found for pelvis width and weight (effect on the translational peak gain range from 6.05 to -0.06). In fore-aft and lateral perturbation, taller participants have larger translation gains than mid and small height participants (Figure 6, b). In contrast, the vertical perturbation has an opposite effect of length (tall = 4.78; mid = 5.17; small = 5.57 units).

A modest interaction effect was observed between the adopted sitting posture and body weight. During trials with upright active sitting posture participants with lower body weight could manage to reduce the translational gain (the effect ranged between 2.63 to 4.01 units for the erect posture). Having more extensive thorax-pelvis angles leads to a larger body kinematic response (preferred from 3.02 to 3.65; slouched from 2.99 to 3.84).

Typically, the pelvis and trunk gains increased from lateral to fore-aft to vertical perturbation (pelvis from 1.30 to 1.88 to 2.83, and trunk from 2.76 to 3.09 to 4.44, for fore-aft, lateral and vertical). The lateral perturbation increased head translational gain (3.97 units) compared to fore-aft (2.07 units). In the vertical perturbation the head kinematic responses were the strongest (effect of 5.46).

## IV. DISCUSSION

The present study investigates the predictability of kinematic peak gains using multiple linear regression models for fore-aft, lateral, and vertical perturbations. By taking into account motion direction and body segment, over 72% of the peak translational gains could be explained. Adding adopted sitting postures (i.e., slouched, upright passive and active) and biological sex (as a categorical predictor) does not significantly affect the model's coefficient of determination. This indicates that these variables do have much of an effect on body kinematic responses.

To obtain a comprehensive understanding of biodynamic responses in whole-body vibration requires consideration of interactions between vibration direction, seat characteristics, sitting postures, and anthropometric attributes, amongst others [13]. Therefore, we have evaluated these interactions in the stepwise multiple regression procedure. This resulted in a model capable of explaining over 90% of the variance in translational peak gains. This paper shows common trends in head, trunk and pelvis, in all three platform motion conditions in Figs 5&6. In addition, the peak gains were found to be highly dependent on the platform motion direction and on the specific body segment. Similar results can been see at the work presented by Mirakhorlo et al. [6]. The model output indicated that the most effective pelvis and trunk-in-space stabilization was found in the lateral, while the head peak translational gain are reduced in the fore-aft perturbation.

The interaction effects show how effects of a given variable can change depending on the level of another variable. The stepwise regression model revealed that the effect of body weight on translational gains changes direction from positive to negative with increasing body length, similar findings can be found in the literature [11]. Additionally, that the effect of motion direction changes with the length of participants: vertical perturbations affect smaller participants the most, while in the fore-aft and lateral direction, taller participants will be affected the most. Furthermore, the model revealed that having a larger base support (hip width, larger distances between anterior superior iliac) over the seat pan area increases translational gains. [4].

In a real driving scenario, where participants and drivers are exposed to prolonged driving (more than 2 hours), it is expected that they adjust their posture to optimize their perceived comfort [14]. The postural differences will also be affected by other factors (i.e., body height, discomfort, age) [4]. Those posture alternations are expected to influence the transmissibility of whole-body vibration [3], [6]. Our models did show an effect of postural alteration on translational peak gains. Nevertheless, these effects were not as significant as those associated with the other predictors (motion direction, body weight, etc.). Despite considering the interaction effect with other dependent predictors (direction, biological sex-based differences, body segment responses, among others), the models suggest that posture variation weakly influences the outcome. This limited effect of posture on the models' predictions might be related to the relatively narrow study population. Further research with a more diverse sample and a wider range of perturbation bandwidth could provide more insights into the relationship between selected dependent predictors and peak translational gains.

## V. Conclusion

A sample of 35 participants was exposed to random vibration in a tri-axial direction with a moving base driving simulator. To explain peak gains of the kinematic body responses, four (three basic and one exploratory) models were created. By employing these models, we uncovered that perturbation direction and body segment explained much of the variance in translational gains. Model predictions improved by introducing interactions between recorded anthropometric attributes, biological sex, and adopted sitting postures. Despite the expectation that changes in sitting posture would influence peak translational gains, our regression models revealed only a small effect of posture variation on the outcome. Additionally, the effect of biological sex was limited. Basic body attributes (stature and weight) do have a considerable effect on the model's predictions. These findings highlight the complexity of the relationship between adopted sitting posture, measured body attributes, motional direction, and peak translational gains during exposure to random vibration.

## Appendix

TABLE I. EXPLORATORY MODEL. STEPWISE LINEAR REGRESSION MODEL. $R^2 = 0.914$ AND $R^2$ ADJUSTED = 0.907

| Term names | Estimate | SE | tStat | pValue |
|---|---|---|---|---|
| Intercept | -81.36 | 32.30 | -2.52 | 0.012 |
| AG | -0.59 | 0.13 | -4.49 | 0.000 |
| BW | 0.31 | 0.08 | 4.03 | 0.000 |
| BH | 0.68 | 0.24 | 2.80 | 0.005 |
| HH | 0.74 | 0.31 | 2.38 | 0.018 |
| HW | -0.95 | 0.20 | -4.74 | 0.000 |
| TH | -0.28 | 0.26 | -1.07 | 0.283 |
| BS_F | 12.86 | 2.85 | 4.51 | 0.000 |
| Se_Tr | 4.57 | 1.33 | 3.44 | 0.001 |
| Se_H | 0.09 | 1.32 | 0.07 | 0.948 |
| Di_La | -0.05 | 1.45 | -0.04 | 0.971 |
| Di_Ve | 4.13 | 1.48 | 2.79 | 0.005 |
| Po_P | 0.90 | 0.25 | 3.66 | 0.000 |
| Po_Sl | 0.69 | 0.25 | 2.77 | 0.006 |
| A:BW | 0.00 | 0.00 | 2.88 | 0.004 |
| A:BH | 0.01 | 0.00 | 4.40 | 0.000 |
| A:HH | -0.01 | 0.00 | -4.60 | 0.000 |
| A:HW | 0.01 | 0.00 | 5.64 | 0.000 |
| A:TH | 0.00 | 0.00 | -4.04 | 0.000 |
| A:BS_F | 0.08 | 0.01 | 5.90 | 0.000 |
| A:Se_Tr | 0.01 | 0.00 | 1.74 | 0.081 |
| A:Se_H | 0.01 | 0.00 | 3.43 | 0.001 |
| BW:BH | 0.00 | 0.00 | -3.73 | 0.000 |
| BW:HW | 0.00 | 0.00 | -5.07 | 0.000 |
| BW:Se_Tr | 0.03 | 0.00 | 5.91 | 0.000 |
| BW:Se_H | 0.01 | 0.00 | 2.65 | 0.008 |
| BW:Di_La | 0.00 | 0.00 | 0.70 | 0.484 |
| BW:Di_Ve | 0.01 | 0.00 | 3.17 | 0.002 |
| BW:Po_P | -0.01 | 0.00 | -3.06 | 0.002 |
| BW:Po_Sl | -0.01 | 0.00 | -2.22 | 0.027 |
| BG:HH | -0.01 | 0.00 | -2.86 | 0.004 |
| BH:TH | 0.00 | 0.00 | -3.24 | 0.001 |
| BH:Se_Tr | -0.03 | 0.01 | -4.03 | 0.000 |
| BH:Se_H | -0.01 | 0.01 | -0.71 | 0.477 |
| BH:Di_La | -0.01 | 0.02 | -0.54 | 0.587 |
| BH:Di_Ve | -0.07 | 0.02 | -4.10 | 0.000 |
| HH:HW | 0.01 | 0.00 | 5.02 | 0.000 |
| HH:TH | 0.01 | 0.00 | 3.05 | 0.002 |
| HH:BS_F | -0.10 | 0.02 | -4.43 | 0.000 |
| HH:Di_La | 0.01 | 0.02 | 0.27 | 0.790 |
| HH:Di_Ve | 0.05 | 0.02 | 2.50 | 0.012 |
| TH:BS_F | -0.13 | 0.03 | -4.25 | 0.000 |
| TH:Di_La | 0.00 | 0.01 | 0.34 | 0.736 |
| TH:Di_Ve | 0.05 | 0.01 | 3.42 | 0.001 |
| BS_F:Se_T | 0.37 | 0.13 | 2.72 | 0.007 |
| BS_F:Se_H | -0.11 | 0.13 | -0.80 | 0.425 |
| BS_F:Di_La | -0.03 | 0.14 | -0.21 | 0.832 |
| BS_F:Di_VE | 0.33 | 0.14 | 2.40 | 0.017 |
| Se_Tr:Di_La | 0.25 | 0.11 | 2.27 | 0.024 |
| Se_H:Di_La | 0.40 | 0.11 | 3.64 | 0.000 |
| Se_Tr:Di_Ve | 2.48 | 0.11 | 22.01 | 0.000 |
| Se_H:Di_Ve | 2.44 | 0.11 | 22.15 | 0.000 |
| Se_Tr:Po_P | -0.09 | 0.11 | -0.79 | 0.432 |
| Se_H:Po_P | -0.15 | 0.11 | -1.37 | 0.172 |
| Se_Tr:Po_Sl | -0.38 | 0.11 | -3.45 | 0.001 |
| Se_H:Po_Sl | -0.49 | 0.11 | -4.44 | 0.000 |
| Di_La:Po_P | -0.04 | 0.11 | -0.41 | 0.685 |
| Di_Ve:Po_P | 0.03 | 0.11 | 0.30 | 0.762 |
| Di_La:Po_Sl | 0.22 | 0.11 | 2.01 | 0.045 |
| Di_Ve:Po_Sl | 0.63 | 0.11 | 5.58 | 0.000 |

A- Age; BW – Body Wight; BH – Body Height; HH – hip height; HW – hip width; TH – Trunk Height; BS_F – Biological sex Female; SE_Tr – Segment Trunk; SE_H – Segment Head; Di_La – Direction Lateral; Di_Ve – Direction Vertical; Po_P – Posture Preferred; Po_Sl – Posture Slouched.

## Acknowledgment

We acknowledge the support of Toyota Motor Corporation in funding the research.

## References

[1] N. Nawayseh, A. Alchakouch, and S. Hamdan, "Tri-axial transmissibility to the head and spine of seated human subjects exposed to fore-and-aft whole-body vibration," *J. Biomech.*, vol. 109, p. 109927, 2020, doi: 10.1016/j.jbiomech.2020.109927.

[2] M. J. Griffin and J. Erdreich, "Handbook of human vibration." Acoustical Society of America, 1991.

[3] B. Basri and M. J. Griffin, "Equivalent comfort contours for vertical seat vibration: Effect of vibration magnitude and backrest inclination," *Ergonomics*, vol. 55, no. 8, pp. 909–922, 2012, doi: 10.1080/00140139.2012.678390.

[4] M. G. R. Toward and M. J. Griffin, "The transmission of vertical vibration through seats: Influence of the characteristics of the human body," *J. Sound Vib.*, vol. 330, no. 26, pp. 6526–6543, 2011, doi: 10.1016/j.jsv.2011.07.033.

[5] B. Basri and M. J. Griffin, "Predicting discomfort from whole-body vertical vibration when sitting with an inclined backrest," *Appl. Ergon.*, vol. 44, no. 3, pp. 423–434, 2013, doi: 10.1016/j.apergo.2012.10.006.

[6] M. Mirakhorlo, N. Kluft, B. Shyrokau, and R. Happee, "Effects of seat back height and posture on 3D vibration transmission to pelvis, trunk and head," *Int. J. Ind. Ergon.*, vol. 91, no. May, p. 103327, 2022, doi: 10.1016/j.ergon.2022.103327.

[7] M. Demić and J. Lukić, "Investigation of the transmission of fore and aft vibration through the human body," *Appl. Ergon.*, vol. 40, no. 4, pp. 622–629, 2009.

[8] N. Nawayseh, "Transmission of vibration from a vibrating plate to the head of standing people," *Sport. Biomech.*, vol. 18, no. 5, pp. 482–500, 2019.

[9] R. H. R. Desai, M. Cvetkovic, G Papaioannou, "Computationally efficient human body modelling for real time motion comfort assessment."

[10] M. Mirakhorlo *et al.*, "Simulating 3D Human Postural Stabilization in Vibration and Dynamic Driving," *Appl. Sci.*, vol. 12, no. 13, 2022, doi: 10.3390/app12136657.

[11] M. G. R. Toward and M. J. Griffin, "Apparent mass of the human body in the vertical direction: Inter-subject variability," *J. Sound Vib.*, vol. 330, no. 4, pp. 827–841, 2011, doi: 10.1016/j.jsv.2010.08.041.

[12] M. M. Cvetkovic, D. Soares, and J. S. Baptista, "Assessing post-driving discomfort and its influence on gait patterns," *Sensors*, vol. 21, no. 24, p. 8492, 2021.

[13] S. Rakheja, K. N. Dewangan, R. G. Dong, and P. Marcotte, "Whole-body vibration biodynamics-a critical review: I. Experimental biodynamics," *Int. J. Veh. Perform.*, vol. 6, no. 1, pp. 1–51, 2020.

[14] M. Cvetkovic, D. Soares, P. Fonseca, S. Ferreira, and J. S. Baptista, "Changes in postures of male drivers caused by long-time driving," in *Occupational and Environmental Safety and Health II*, Springer, 2020, pp. 491–498.

[15] K. N. de Winkel, T. Irmak, V. Kotian, D. M. Pool, and R. Happee, "Relating individual motion sickness levels to subjective discomfort ratings," *Exp. Brain Res.*, vol. 240, no. 4, pp. 1231–1240, 2022, doi: 10.1007/s00221-022-06334-6.